\documentclass[10pt,onecolumn,letterpaper]{article}

\usepackage{times}
\usepackage{epsfig}
\usepackage{graphicx}
\usepackage{amsmath}
\usepackage{amssymb}
\usepackage{array}
\usepackage{adjustbox}
\usepackage{array}
\usepackage{cellspace}
\usepackage{hhline}

\graphicspath{{figuras/}}
\usepackage[utf8]{inputenc} 
\usepackage{multirow}
\usepackage{placeins}
\usepackage{lineno}
\usepackage{color}
\usepackage{hyperref}
\usepackage{subcaption}
\pdfoutput=1

\definecolor{greenao}{rgb}{0.0, 0.5, 0.0}

\newcommand*{\mathcolor}{}
\def\mathcolor#1#{\mathcoloraux{#1}}
\newcommand*{\mathcoloraux}[3]{  \protect\leavevmode
	\begingroup
	\color#1{#2}#3  \endgroup
}

\definecolor{111}{rgb}{0,1,0}
\definecolor{00}{rgb}{1.00000000000000000000,.49803921568627450980,0}
\definecolor{222}{rgb}{0.8,0.9568627451,0.56470588235}
\definecolor{111}{rgb}{0.8,0.9568627451,0.56470588235}

\usepackage[table]{xcolor}

\begin{document}
\title{3DFCNN: Real-Time Action Recognition using 3D Deep Neural Networks with Raw Depth Information}

\author{Adrian Sanchez-Caballero,  Sergio de López-Diz,\\
	 David Fuentes-Jimenez, Cristina Losada-Gutiérrez,\\
	  Marta Marrón-Romera, David Casillas-Perez, Mohammad Ibrahim Sarker\\
Universidad de Alcal\'a\\
{\tt\small \{adrian.sanchez,s.lopezd,d.fuentes, david.casillas\}@edu.uah.es}\\
{\tt\small\{marta.marron, cristina.losada,ibrahim.sarker\}@uah.es} }
\maketitle

\begin{abstract}
Human actions recognition is a fundamental task in artificial vision, that has earned a great importance in recent years due to its multiple applications in different areas. 
In this context, this paper describes an approach for real-time human action recognition from raw depth image-sequences, provided by an RGB-D camera. The proposal is based on a 3D fully convolutional neural network, named 3DFCNN, which automatically encodes spatio-temporal patterns from depth sequences without 
pre-processing. Furthermore, the described 3D-CNN allows 
actions classification from the spatial and temporal encoded information of depth sequences. The use of depth data ensures that action recognition is carried out protecting people's privacy
, since their identities can not be recognized from these data. 
3DFCNN has been evaluated and its results compared to those from other state-of-the-art methods within three widely used 
datasets, with different characteristics (resolution, sensor type, number of views, camera location, etc.). The obtained results allows validating the proposal, concluding that it outperforms several state-of-the-art approaches based on classical computer vision techniques. Furthermore, it achieves action recognition accuracy comparable to deep learning based state-of-the-art methods with a lower computational cost, which allows its use in real-time applications.
\end{abstract}

\section{Introduction}
\label{sec:introduction}

In the context of artificial vision, human action recognition (HAR) has gained a great importance in recent years, mainly due to its multiple applications in the study of human behavior, safety or video surveillance, which has attracted the attention of many researchers~\cite{poppe2010,weinland2011,chen2017,ko2018,Wang2020}.

There are several works in the literature whose aim is recognizing human actions. The first proposals were based on the analysis of RGB sequences~\cite{Tafazzoli2010,sadanand2012,Wang2013ICCV,zhang2017}, using different datasets~\cite{chaquet2013}. The release of RGB-D cameras~\cite{han2013,Sell2014}, that in addition to a color image provide a depth map (in which each pixel represents the distance from the corresponding point to the camera), has allowed the appearance of numerous works that address HAR using RGB-D information~\cite{ashraf2014,LIU201574,farooq2015survey,KHAIRE2018107,al2018human} or 3D skeleton data~\cite{KHAIRE2018107, weng2017spatio,laraba20173d} with acceptable results. In addition, several new datasets including RGB-D information for action recognition have been made available to the scientific community~\cite{zhang2016}. 

Most of the previously cited proposals provide good results in controlled conditions, however, they present problems in scenarios with a high degree of occlusions. Besides, the use of color information (RGB or RGB-D) implies the existence of data that allows the users' identification, so problems related to privacy may appear. The use of depth cameras~\cite{Lange2001}, that obtain just information about the distance from each point of the scene to the camera by indirectly measuring the time of flight of a modulated infrared signal, allows to preserve people privacy
Another advantage to consider is that these cameras do not require additional lighting sources, as they include an infrared lighting source. Thus, depth maps are also widely used in different works for action recognition~\cite{wang2015convnets,wang2015deep,wang2015action,wu2019hierarchical}.  

In recent years, the improvements in technology and the appearance of large-scale datasets have led to an increase in the number of HAR works based on deep learning using 
RGB sequences~\cite{ji20123d,wang2016temporal,feichtenhofer2016}, 
RGB-D data~\cite{wang2018survey,hu2018ECCV,das2019}, depth maps~\cite{luo2017unsupervised,wang2018depth,xiao2019action} 
or 3D skeletons~\cite{ke2017new,li2017skeleton,kim2017interpretable,KHAIRE2018107}. 
All these works provide good accuracy but with a high computational cost, which mostly prevents its operation in real-time applications.

Despite the numerous works dealing with the recognition of actions, this is then still an open issue in real scenarios, with open problems such as the different viewpoints in videos, changing lighting conditions, occlusions, etc. 

In this context, this paper describes an approach for real-time action recognition, which performs the features extraction and action classification steps in a single stage, through the use of 3D Convolutional Neural Networks (3D-CNNs). The proposal is based on a 3D fully convolutional neural network, named 3DFCNN, and uses only the raw depth information provided by a depth or RGB-D camera. It is an end-to-end trainable model, composed by a first phase to extract the main spatial and temporal features, using 3D convolutions and pooling layers, and a final softmax layer for obtaining the detected action. It is worth highlighting that, although the proposal is based on deep learning, the network architecture has been optimized so it is able to work in real-time. 

Both training and testing stages have been carried out with the widely used ``NTU RGB+D Action Recognition Dataset''~\cite{Shahroudy_2016_CVPR,Liu_2019_NTURGBD120}, made available to the scientific community by the ROSE Lab of the Nanyang Technological University of Singapore. This dataset has been chosen because it provides a large number of videos, both with RGB and depth information, and it allows to compare, through the experimental results, different works of the state-of-the-art. Furthermore, to test the robustness of the 3DFCNN and compare it to that of other previous works, it has also been evaluated 
using two multiview 3D action datasets: NorthWest-UCLA Multiview Action 3D~\cite{nwucla} and UWA3D Multiview Activity II~\cite{uwa3dii}.

The rest of this paper is organized as follows: in section~\ref{sec:related-works} the main related works are presented and analyzed, next, in section~\ref{sec:architecture} the architecture of the neural network proposed is described. Then, in section~\ref{sec:training}, the training method used is explained. Subsequently, section~\ref{sec:results} includes the main experimental results obtained, and finally, section~\ref{sec:conclusions} includes the main conclusions of the work, as well as some possible lines of future work.

\section{Related works}
\label{sec:related-works}

As stated in the introduction, multiple proposals for HAR have been developed by the scientific community during the last decade, based on different visual technologies. In this section, the most interesting ones are analyzed, in order to be compared with the proposal hereby described. The main works related to HAR can be divided into three groups depending on the technology used: RGB, RGB-D and depth images~\cite{Singh2019,zhang2019comprehensive}. Below, it is presented a brief analysis of the RGB and RGBD-based works, and a more in-depth study of those using only depth.

The first works in HAR were based on the use of RGB sequences~\cite{poppe2010,sadanand2012,Baptista2016,chou2018robust}. These works require a feature extraction system followed by a classification process for action recognition. More recently, the improvements in technology and the availability of large-scale datasets have led to an increase in the number of related works that use RGB data and are based on deep learning techniques~\cite{chahramani2014,wang2016temporal,feichtenhofer2016,wang2018human}.

When combining the RGB channels of an image with its depth information, RGB-D images are obtained. In recent times, analysis of these images or videos are sought due to the availability of real-time inexpensive depth sensors that provide rich 3D structural data. 
Thus, this has motivated the proposal of numerous works based on combining RGB and depth data for HAR~\cite{zhao2012,Hu2015CVPR,LIU201574,farooq2015survey,hsu2016,liu2019rgbd}.

Also, the promising results achieved by deep learning methods in computer vision applications have encouraged their utilization on RGB-D images and videos for HAR~\cite{khurana2018deep,wang2018survey,liu2018viewpoint,hu2018ECCV,wang2019generative,kong2019,das2019}, mostly based on the use of CNNs~\cite{al2018human} and Recurrent Neural Networks (RNNs). 

Besides, regarding to HAR proposals that rely on just depth information, also several modalities of depth data have been used in the related literature. Some studies use raw depth maps~\cite{wang2015convnets,wang2015deep,wang2015action,luo2017unsupervised, wang2018depth,wu2019hierarchical,xiao2019action}, whereas others extract specific 3D information from them, as a pre-processing task, like joint positions of human body (skeleton data)~\cite{song2017end,liu2016spatio,ke2017new,zhang2017view,li2018independently,yan2018spatial,si2019attention}. In both cases, deep learning methods have replaced conventional methods to process this data, especially when a large scale dataset is available.  

The main contributions to action recognition using 3D skeletons have been focused on input data representations and improvements of deep learning methods. Although there have been some research proposals for action recognition based on CNN using 3D skeleton data with good results~\cite{ke2017new, li2017skeleton, kim2017interpretable}, most recent studies use RNNs and, in particular, variations of the long short-term memory (LSTM) unit, which solves the gradient vanishing and exploding problems~\cite{greff2016lstm, jozefowicz2015empirical}. 
In addition, LSTM networks are able to learn long-term dependencies, which is essential in action recognition problems. 
Song \emph{et al.} proposed in~\cite{song2017end} a modified LSTM with a joint selection gate for creating an spatio-temporal attention model. In a similar way, in~\cite{liu2016spatio} it was introduced a trust gate for the LSTM and a 3D skeleton data representation in a tree-like graph was proposed, whereas other proposals built a view adaptive LSTM scheme~\cite{zhang2017view} to solve the problem of sensitive viewpoint variations in human action videos. 

Li \emph{et al.} in~\cite{li2018independently} presented a new type of RNN: the independently RNN (IndRNN), that improved previous results in 3D skeleton-based action recognition. In IndRNN, neurons inside a layer are independent among each other but connected across layers, favouring the stacked-layers scheme. 

Recent approaches use an extension of the graph convolutional networks (GCN) in order to allow the network learning spatio-temporal features more efficiently. Yan \emph{et al.} in~\cite{yan2018spatial} proposed a spatio-temporal graph convolutional network (ST-GCN) which automatically learns both the spatial and temporal patterns from 3D skeleton data by, firstly, transforming it into a graph. 

Similarly, Si \emph{et al.}, in~\cite{si2019attention}, achieved state of the art results by using an attention enhanced graph convolutional LSTM network (AGC-LSTM). This model presents also a temporal hierarchical architecture and takes into account the co-ocurrence relationship between spatial and temporal domains. 

Besides, other researchers have directly used depth maps, avoiding some known problems related to 3D skeleton position extraction as spatial information loss in images, extraction failures and sensitivity to pose variations. Moreover, by using raw depth maps, the entire 3D structure information of a scene can be used for recognition. The similarity of depth images with RGB ones, makes it possible to transfer all the knowledge from RGB-based action recognition proposals to the depth modality. Thus, the success of CNN methods in RGB-based recognition methods makes reasonable to transfer these studied techniques to the depth domain, and that is what most recent studies have done. For instance, most works~\cite{wang2015convnets,wang2015deep,wang2015action,wang2018depth} have used a pseudocoloring technique by which a depth sequence is encoded into several RGB images (depth motion maps and dynamic images), transforming spatio-temporal patterns to colored properties like textures, as it is explained next. 

In addition, some approaches as the ones described in~\cite{wang2015convnets,wang2015deep,wang2015action} apply 3D rotations to point clouds from depth images, to use 3 orthogonal projections, leveraging the 3D structure of depth data and, concurrently, augmenting samples of the training dataset.
In these approaches, depth motion maps are generated from these projected sequences using different temporal pooling methods, referred to as rank pooling. Alternatively, Luo \emph{et al.} in~\cite{luo2017unsupervised} proposed an encoder-decoder framework using an unsupervised LSTM network through 3D flows for motion description. 

Similar to depth motion maps, three sets of dynamic images (DI) were proposed in~\cite{wang2016large,wang2018depth} for encoding depth sequences: dynamic depth image (DDI), dynamic depth normal image (DDNI) and dynamic depth motion normal image (DDMNI). These images are constructed using hierarchical and bidirectional rank pooling to capture spatio-temporal features. Therefore, for each video sample, it is used an ensemble of 6 images as input of 6 independent CNNs through a VGG-16 architecture~\cite{simonyan2014deep}. 

Also, spatially structured dynamic depth images have been used for selecting attributes to be used for action recognition \cite{wang2017structured}. In this approach, rank pooling has been employed to extract three pairs of structured dynamic images, one at each body part, and with joint level granularity. This aids in retaining the spatio-temporal details as well as in improving the structural particulars at different granularity at various temporal scales. Further, spatially and temporally structured dynamic depth images have been highlighted through hierarchical and bidirectional rank pooling methods in order to derive spatial, temporal and structural characteristics from the depth channel of the image \cite{hou2017spatially}.

In a similar way, Wu \emph{et al.} in~\cite{wu2019hierarchical} proposed using depth projected difference images as a dynamic image with a hierarchical rank pooling with 3 pre-trained CNNs, one for each orthogonal projection. 

Finally, a multi-view imaging framework was proposed in~\cite{xiao2019action} for action recognition with depth sequences. First, it is employed a human detection network to reduce the action region. A total of 11 different viewpoints are used then for generating projected images at several temporal scales, making thus data augmentation. Also, they propose a novel CNN model where dynamic images with different viewpoints share convolutional layers but have different fully connected layers for each view. The final action classification is made with support vector machines (SVM) with principal component analysis (PCA). All this process results in a method that achieves state of the art on action recognition accuracy on several datasets in exchange of a high computational cost.

\section{Network Architecture}
\label{sec:architecture} 

\begin{figure}[htbp]
	\centering
	\includegraphics[width=0.8\textwidth]{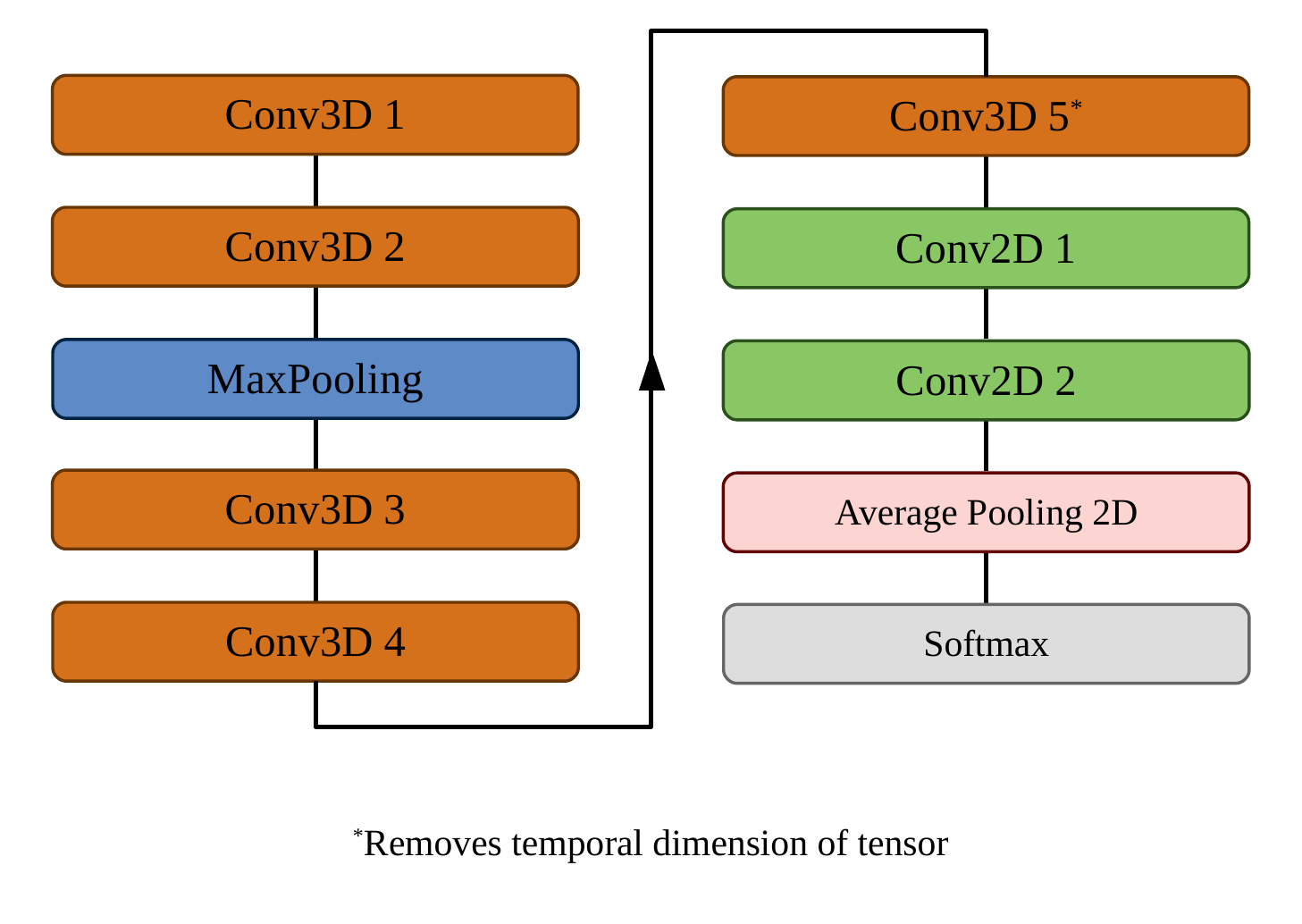}
	\caption{Simplified architecture summary of the proposed 3DFCNN.}
	\label{fig_arquitectura}
\end{figure}

We propose an end-to-end trainable deep learning model for the action recognition problem with depth maps named 3DFCNN. It is a 3D fully CNN. 
This architecture has been chosen because 3DCNNs have demonstrated their efficacy for action recognition in RGB videos
~\cite{dipakkr}. 
Next, it is described in detail the complete architecture of the used neural network, whose general structure is shown in Fig.~\ref{fig_arquitectura}. 

The network input is a sequence of depth images, sized $64 \times 64$ pixels, with a fixed number of frames. Specifically, it is a 1 second (30 frames) video fragment, showing the execution of a single action.

This number of frames has been chosen to achieve a balance between accuracy and computation time. 
The use of sequences with a reduced number of frames allows decreasing the computational cost and enables real-time performance. Besides, in some works such as in~\cite{schindler2008action} 7-10 frames are enough for recognizing simple actions. However, in the case of NTU RGB+D dataset~\cite{Shahroudy_2016_CVPR}  that includes complex and 
very long actions (the maximum video length is 300 frames), 
this sequence length may not be enough to fully cover them. 
In contrast, the use of longer sequences increases the computational cost, reducing the processing speed. Moreover, in this situation, overlapping may occur between different actions in continuous action videos.

In this paper, an experimental adjustment of the number of frames has been made, 
with the aim of achieving a balance between the computational cost 
and the accuracy of the algorithm. The optimal length has been determined to be of 1 second (30 frames) for the input sequences of our deep learning model. It is enough for including significant information for most of the detected actions, without increasing the computation cost. Besides, it reduces the possibility of several actions overlapping in the same video-segment in real sequences, which can include more than one action throughout the video.
Since the NTU RGB+D dataset includes videos with very different length, there has been proposed a method for selecting 30 frames for each complete video, that is detailed later, in section~\ref{sec:training}.



\begin{figure}[htbp]
	\centering
	\includegraphics[width=0.6\textwidth]{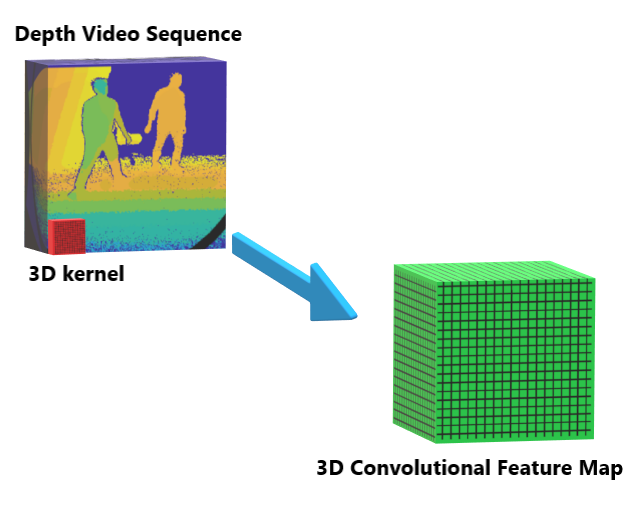}
	\caption{Example of a 3D convolutional operation on a colored video depth sequence applying a 3D kernel (red), and result of the operation (green).}
	\label{fig_conv3d}
\end{figure}

In contrast to 2D CNNs, in which the operations are carried out only on the spatial dimension of input images, in a 3D-CNN, features are extracted by applying 3D convolutional filters on the spatial and temporal dimensions of input videos. This is necessary for the neural network in order to take into account the context and temporal changes of actions for a better recognition. Fig.~\ref{fig_conv3d} shows an illustrative example of the 3D convolution operation and the appearance of the applied kernel. 

Input sequences are processed by first applying two 3D convolutional layers with 32 filters each, padding with zeroes in order to conserve tensor dimensions, and then a dimensionality reduction through a \emph{Max Pooling} layer. Second, another convolutional block of two layers with 64 filters each one, without any padding. Next, the temporal dimensionality is removed with an additional Conv3D layer with 128 non-squared filters. After that, a Conv2D layer with 128 filters precedes a final Conv2D layer with a number of filters that matches classes number in the dataset (60 in case of NTU RGB+D dataset).

Finally, an activation layer ``softmax", noted as $S$, is used to compress the output vector to real values between $0$ and $1$, in order to obtain a normalized likelihood distribution, see equation~\ref{eqn:softmax1}. The expression~\ref{eqn:softmax2} shows the formula applied to obtain the probabilities for each action.


\begin{align}
S \colon \mathbb{R}^{60} & \to [0,1]^{60} \nonumber \\ 
a=\left(a_1,\cdots,a_{60}\right) & \mapsto S(a)=\left(S_1,\cdots,S_{60}\right)
\label{eqn:softmax1}  
\end{align}	

\begin{equation}
S_j=\frac{e^{a_j}}{\sum_{k=1}^{60}e^{a_k}}\ \ \ \ 1\leq j \leq60
\label{eqn:softmax2}
\end{equation}

Besides, at the output of each convolutional layer (except for \emph{Conv3D 5} and \emph{Conv2D 2}), a \emph{Batch Normalization}~\cite{ioffe2015batch} layer and a \emph{LeakyReLU} (\textit{Leaky Rectified Linear Unit}~\cite{maas2013rectifier}) activation function 
are included. Batch Normalization helps training the neural network reducing the internal covariate shift. The Leaky ReLU activation function follows the expression $f(x)=x$ if $x\geq0$ and $f(x)=\alpha x$ if $x<0$, with $\alpha=0.3$. This type of function has been chosen instead of the conventional ReLU due to its proven greater efficiency~\cite{xu2015empirical}, and because it provides the necessary non-linearity to solve the action recognition problem avoiding gradient vanishing problems~\cite{imagenet2012}.

In addition, it is also applied a slight dropout regularization technique for helping the model to generalize more and reduce over-fitting problems during training. 

A more detailed description of the different layers that form the proposed network is shown in Table~\ref{Tabla:1}, as well as their output sizes and fundamental parameters.

\begin{table}[htbp]
	\caption{Network architecture parameters and output tensor sizes for each layer.}
	\label{Tabla:1}
	\centering
		\begin{tabular}{| l | c | c | }
		\hline
		\multicolumn{3}{|c|}{\textbf{Proposed 3D fully convolutional neural network (3DFCNN)}} \\
		\hline
		\hline
		\multicolumn{1}{|c|}{\textbf{Layer}} & \textbf{Output shape} & \textbf{Parameters} \\
		\hline
		\hline
		Input & $64\times64\times30\times1$ & - \\
		\hline
		Conv3D 1 & $64\times64\times30\times32$ & kernel=(3, 3, 3) / strides=(1, 1, 1) \\
		\hline
		Batch Normalization & \multicolumn{2}{c|}{-} \\
		\hline
		Activation & \multicolumn{2}{c|}{LeakyReLU} \\
		\hline
		Conv3D 2 & $64\times64\times30\times32$ & kernel=(3, 3, 3) / strides=(1, 1, 1) \\
		\hline
		Batch Normalization & \multicolumn{2}{c|}{-} \\
		\hline
		Activation & \multicolumn{2}{c|}{LeakyReLU} \\
		\hline
		MaxPooling & $22\times22\times10\times32$ & size=(3, 3, 3) \\
		\hline
		Dropout & \multicolumn{2}{c|}{0.25} \\
		\hline
		Conv3D 3 & $20\times20\times8\times64$ & kernel=(3, 3, 3) / strides=(1, 1, 1) \\
		\hline
		Batch Normalization & \multicolumn{2}{c|}{-} \\
		\hline
		Activation & \multicolumn{2}{c|}{LeakyReLU} \\
		\hline
		Conv3D 4 & $18\times18\times6\times64$ & kernel=(3, 3, 3) / strides=(1, 1, 1) \\
		\hline
		Batch Normalization & \multicolumn{2}{c|}{-} \\
		\hline
		Activation & \multicolumn{2}{c|}{LeakyReLU} \\
		\hline
		Dropout & \multicolumn{2}{c|}{0.25} \\
		\hline
		Conv3D 5 & $18\times18\times1\times128$ & kernel=(1, 1, 6) / strides=(1, 1, 1) \\
		\hline
		Reshape & $18\times18\times128$ & - \\
		\hline
		Conv2D 1 & $8\times8\times128$ & kernel=(3, 3) / strides=(2, 2) \\
		\hline
		Batch Normalization & \multicolumn{2}{c|}{-} \\
		\hline
		Activation & \multicolumn{2}{c|}{LeakyReLU} \\
		\hline
		Conv2D 2 & $8\times8\times60$ & kernel=(1, 1) / strides=(1, 1) \\
		\hline
		Average Pooling 2D & 60 & - \\
		\hline
		Activation & \multicolumn{2}{c|}{softmax} \\
		\hline
	\end{tabular}
\end{table}

The efficient network architecture and the input data generation system make possible to reduce the overall computational complexity of the model, allowing a quite fast performance unlike most other action recognition works in the literature. It opens the door to the capability of real-time performance~\cite{chen2016real,zhang2016real,dawar2017real,liu2018t} and its important applications in video-based health care service, video surveillance or human-computing interaction. 

\section{Network training}
\label{sec:training}

\begin{figure}[!htbp]
	\centering
	\includegraphics[width=0.8\linewidth]{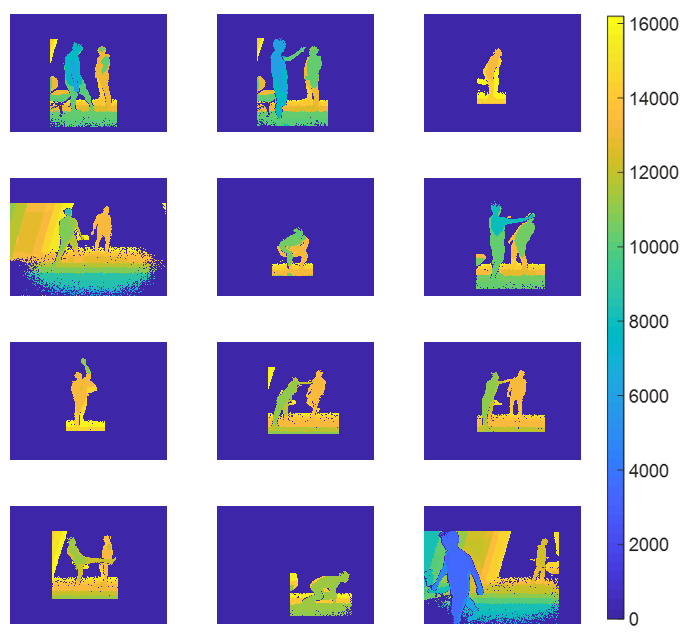}
	\caption{Examples of masked depth images from different video sequences in NTU RGB+D dataset. As shown, it includes 
		several subjects performing different 
		actions from varying viewpoints.}
	\label{fig_dataset}
\end{figure}

The large-scale NTU RGB+D dataset~\cite{Shahroudy_2016_CVPR} has been used for training and testing our model. This dataset contains 56\,880 video sequences for 60 different human actions. These sequences were recorded using \textit{Microsoft Kinect II} sensors ~\cite{zhang2012microsoft} obtaining RGB images, depth maps, 3D skeleton positions and infrared data. In this paper, only depth map videos are used, which are composed of frames with an original resolution of $512\times424$ pixels. However, authors of this dataset suggest to use the masked depth maps, which are the foreground masked version of the full depth mask and, therefore, have a much better compression ratio, which facilitates the download and file managing. Some examples of various masked frames are shown in Fig.~\ref{fig_dataset}.

\begin{figure}[!htbp]
	\centering
	\includegraphics[width=0.7\linewidth]{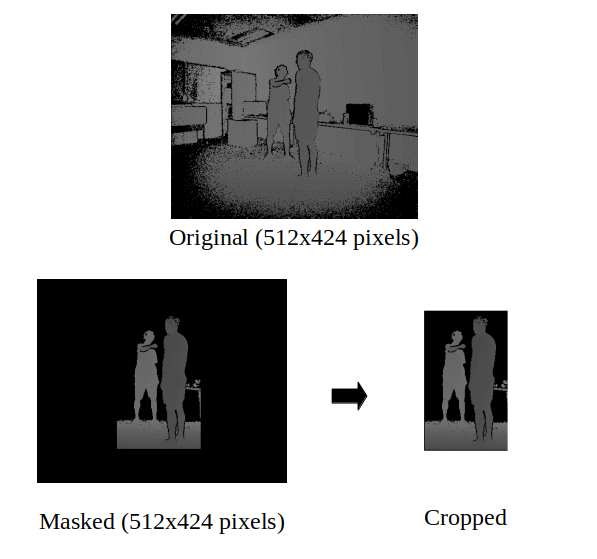}
	\caption{Example of a full depth map (up) and the process of cropping a masked depth map into a smaller image (down).}
	\label{fig_cropping}
\end{figure}

Furthermore, in order to be able to focus on the HAR issue (without previously detecting people) and reduce the number of pixels without useful information, all images have been cropped to the region of interest in which the action is happening, decreasing the weight of files at the same time.
Fig.~\ref{fig_cropping} is an illustrative example of this image cropping. Finally, the cropped image is re-scaled by the model to fit the $64\times64$ input size. 

As it has been explained before, the input of the neural network is composed of 30 frames sequences, so the full input size is $30\times64\times64$ pixels. Experiments have confirmed the importance of how these 30 frames are selected. Consequently, a data arrangement strategy is set so as to take into account the dataset properties used for training. NTU RGB+D dataset contains videos with lengths between 26 to 300 frames. The data generator is specifically modified to optimally select the 30 frames for each sample as follows. When videos are shorter than 30 frames, last frames are repeated backwards until completion. When greater than 30 but shorter than 60 frames, the starting frame is randomly selected inside suitable limits. Finally, when videos are greater or equal to 60 frames the generator randomly selects a suitable starting frame with one-frame-skipping, i.e. taking only odd or even frames. This has two advantages for the network: first, processing the double video length with only 30 frames in case of long actions and second, making data augmentation thanks to random starting point.  

Batch size has been experimentally set to 12, being a reasonable value that makes a good balance between hardware-memory and model generalization. Once the batch size is set, a learning rate range test is performed in order to find which are the best values for this hyperparameter. This test consists in modifying the learning rate value along a wide range and analyzing the loss function behaviour. 

\begin{figure}[!htbp]
	\centering
	\includegraphics[width=0.7\linewidth]{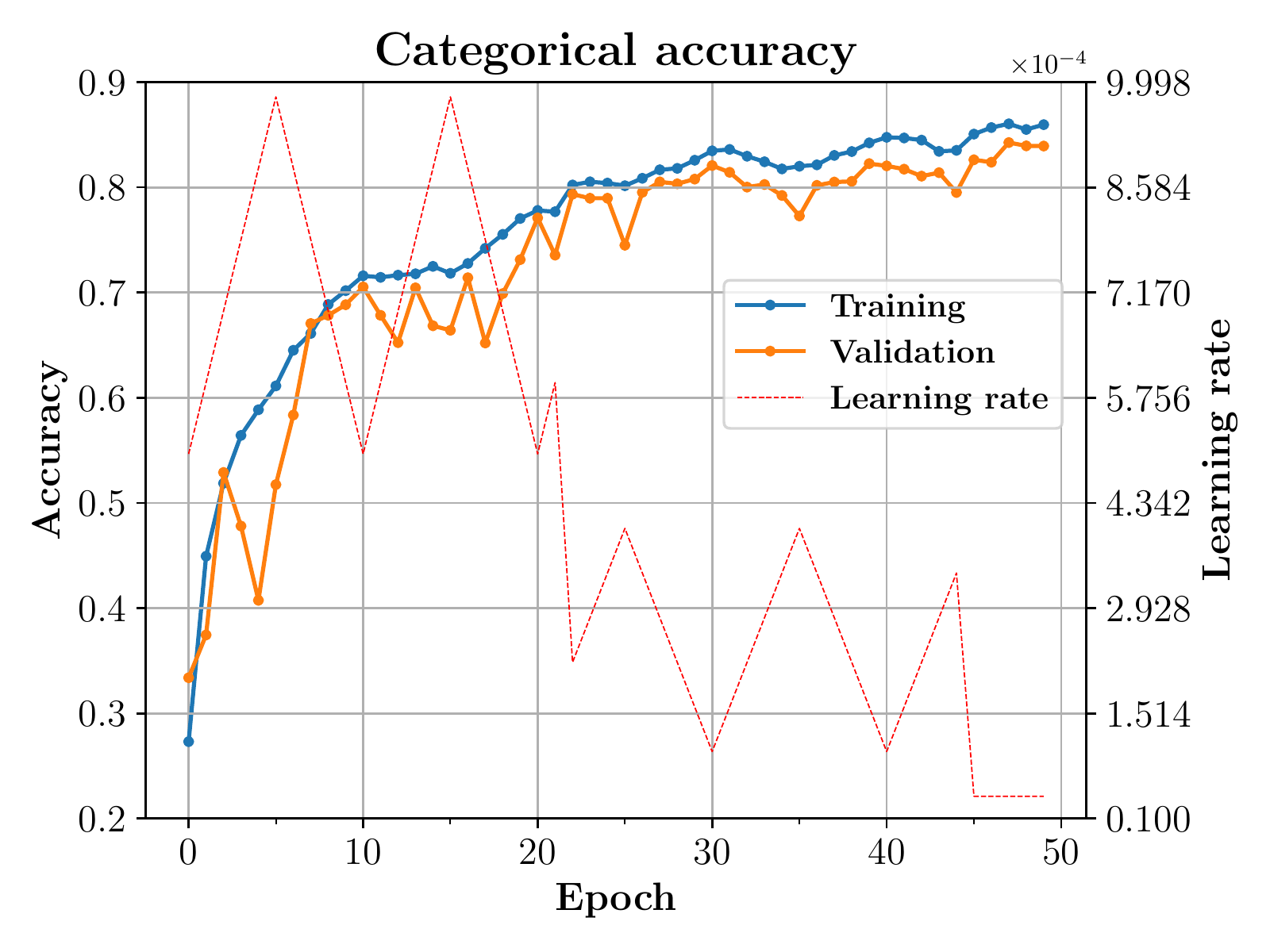}
	\caption{Training and validation categorical accuracy curves. In addition, the learning rate schedule is superposed in red.}
	\label{fig_accuracy1}
\end{figure}

\begin{figure}[!htbp]
	\centering
	\includegraphics[width=0.7\linewidth]{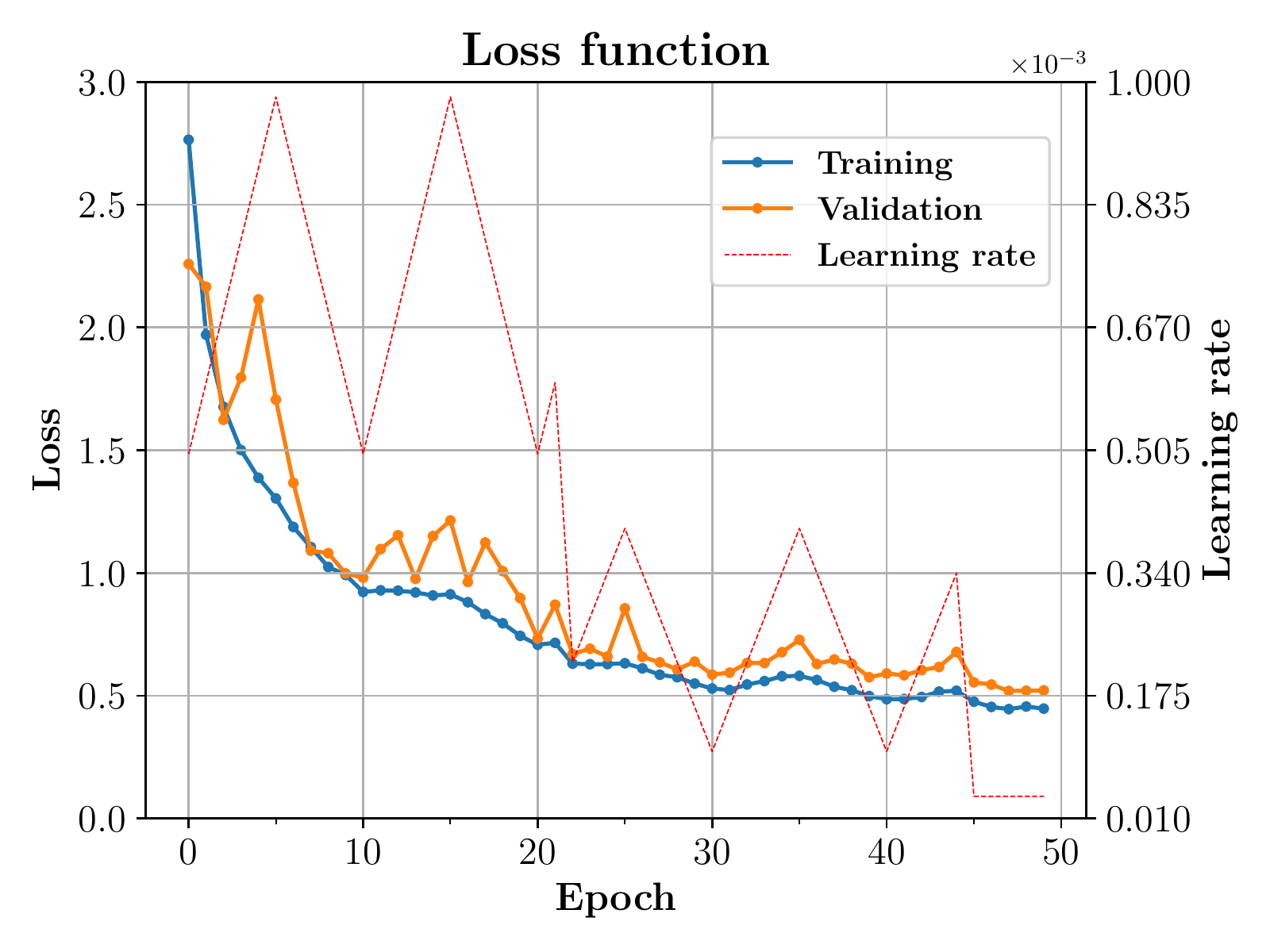}
	\caption{Training and validation loss function curves. In addition, the learning rate schedule is superposed in red.}
	\label{fig_accuracy2}
\end{figure}

A cyclical learning rate schedule is chosen due to its benefits~\cite{smith2017cyclical}, like a faster convergence and a reduced over-fitting. Taking into account the range test results, the learning rate is made to oscillate first between $5\times10^{-4}$ and $9.8\times10^{-4}$, then between $1\times10^{-4}$ and $4\times10^{-4}$ and finally fixed to $4\times10^{-5}$ in the last 5 epochs. The specific curves of learning rate along the training can be seen in Fig.~\ref{fig_accuracy1} and Fig.~\ref{fig_accuracy2}, where accuracy and loss functions for training and validation are shown, respectively. For the optimization process, the algorithm \textit{Adam}~\cite{Kingma2014} has been used due to its proven adaptive properties and computational efficiency. The training of the neural network has been carried out on a NVIDIA GeForce GTX 1080 with 8 GB. 

As it has been said before, Fig.~\ref{fig_accuracy1} and Fig.~\ref{fig_accuracy2} show the process of training the neural network along 50 epochs. This number of epochs have demonstrated to be enough for the model to reach convergence. It can be seen that, thanks to the applied deep learning techniques (dropout, fully-convolutional and cyclical learning rate), although a large dataset as NTU RGB+D is used, validation and training curves are significantly close to each other, i.e. no over-fitting appears. Furthermore, it can be observed that the validation loss slightly diverges (or analogously, the validation accuracy falls) when the learning rate takes the high values of the cycle. This is a desired effect because it prevents the solution to stay stuck at saddle points and helps reaching better minima.

\section{Experimental results and discussion}
\label{sec:results}



\subsection{Experimental setup}

The proposal has been trained and tested using NTU dataset~\cite{Shahroudy_2016_CVPR}. This dataset is used in most of the deep learning based approaches for action recognition because the number of available videos, which allows training deep neural networks. Furthermore, the use of this dataset facilitates comparison to other state-of-the-art approaches, mainly based on DNNs. To evaluate the robustness of the 3DFCNN, it has also been tested in other two multiview publicy available datasets, 
with different actions and camera characteristics, as well as a lower number of sequences for training and testing. These three datasets have been selected to ease comparison with other methods, since there are ones of the most used in the scientific literature for action recognition.
The main characteristics of each dataset are briefly described below. Besides, table~\ref{tab:datasets} shows a summary of these characteristics for allowing comparison. 

\begin{itemize}
	
	\item \textbf{Northwesternt-UCLA Multiview Action3D dataset (NWUCLA)} contains multiview RGB, Depth and 3D joints data acquired using three Kinect v1 cameras in a variety of viewpoints. It includes 10 different actions performed by 10 subjects, with a total of 1494 sequences (518 for view 1, 509 for view 2 and 467 for view 3) with different lengths. In the cross-view setting, the authors propose using two views for training and one view for testing. Figure~\ref{fig_nucla_sample} shows three views of a sample image belonging to this dataset. 
	
	\begin{figure}[!htbp]
		\centering
		\includegraphics[width=\linewidth]{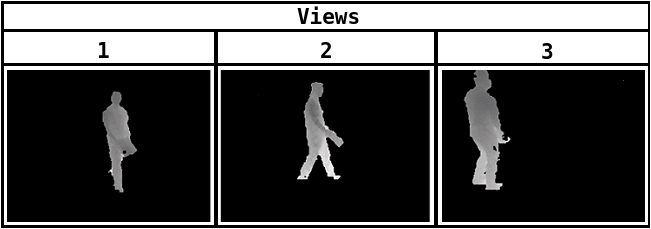}
		\caption{Northwestern-UCLA dataset sample images.}
		\label{fig_nucla_sample}
	\end{figure}
	
	\item {\textbf{UWA3D Multiview Activity II dataset (UWA3DII)}}~\cite{uwa3dii} was collected using a Kinect v1 sensor, and it includes RGB, Depth and 3D joints. This dataset is focused on cross-view action recognition, and includes 10 subjects performing 30 different human activities, recorded from 4 different viewpoints (frontal, left, right and top view). Each subject performs the same action four times in a continuous manner, and each time the camera is moved to record the action from a different viewpoint. As a result, there are 1070 sequences. It is a challenging dataset because of varying viewpoints, self-occlusion and similarity between different actions (such as \textit{walking} and \textit{irregular walking} or \textit{drinking} and \textit{phone answering}).
	The authors propose cross-view action recognition, using the samples from two views for training, and the two remaining views for testing. The complete evaluation includes six different combinations of the 4 views for training and testing, as it can be seen in table~\ref{tab:exp-UWA3DII}. Some sample images of this dataset are shown in figure~\ref{fig_UWA_sample}. 
	
	\begin{figure}[!htbp]
		\centering
		\includegraphics[width=\linewidth]{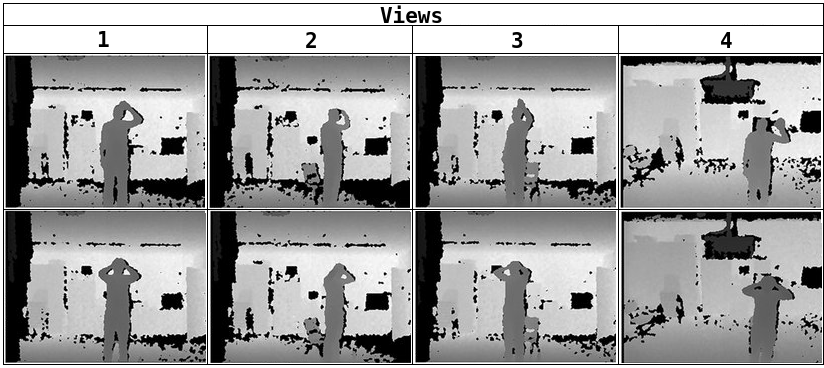}
		\caption{UWA3DII dataset sample images.}
		\label{fig_UWA_sample}
	\end{figure}
	
	\item The large-scale \textbf{NTU RGB+D dataset} contains 56880 video samples for 60 different actions performed by several subjects. The dataset has been acquired using three Kinect V2 cameras concurrently. For each sample, there are RGB videos, depth maps, IR (Infrared) data and 3D joints.The actions in NTU RGB+D dataset can be organized in three major categories: daily actions, mutual actions, and medical conditions. It is worth highlighting that \textit{mutual actions}  (such as \textit{pushing other person}, \textit{hugging other person}, \textit{giving something}, etc.) involve more than one people. Some sample images from this dataset can be seen in figure~\ref{fig_dataset}. 
	The authors of NTU RGB+D dataset propose two different evaluations~\cite{Shahroudy_2016_CVPR} for separating data between training and testing: 
	\begin{enumerate}
		\item \textit{Cross-Subject} (CS) evaluation in which there are 40\,320 training samples with 20 subjects and 16\,560 with other 20 different subjects for testing. 
		\item \textit{Cross-View} (CV) evaluation, with 37\,920 sequences from 2 different viewpoints  for training and 18\,960 from a third camera for testing. 
	\end{enumerate}
	
\end{itemize}

\begin{table}[htbp]
	\centering
	\begin{tabular}{|l|r|r|c|l|m{0.17\textwidth}|}
		\hline
		Dataset & Samples & Classes & Subjects & Sensor & Modalities   \\ \hline \hline
		NWUCLA    &  1494 & 10 & \multirow{2}{*}{10} 
		& \multirow{2}{*}{Kinect v1} 
		& \multirow{2}{0.17\textwidth}{RGB+D, 3DJoints}       \\ \cline{1-3}
		UWA3DII   &  1070 & 30 &    &           &                      \\ \hline
		NTU RGB+D & 56880 & 60 & 40 & Kinect v2 & RGB+D, IR, 3DJoints  \\ \hline 
	\end{tabular}
	\caption{Characteristics of the three datasets used for testing the proposed 3DFCNN. }
	\label{tab:datasets}
\end{table}

As it can be seen in table~\ref{tab:datasets}, the first two datasets (NWUCLA and UWA3DII) include a reduced number of samples, which makes it difficult training DNNs without overfitting. These two datasets also have a lower number of classes than the NTU RGB-D dataset. Besides, the used sensor and the viewpoints are different. Furthermore, it is worth highlighting that the Kinect v1 used in UWA3DII dataset provides images with a higher amount of noise and measurement errors than Kinect v2, as it can be seen in figure~\ref{fig_UWA_sample} 

Due to the small number of available training samples for UWA3DII and NWUCLA datasets, instead of training the network from scratch for each of them, we have started with the model trained using NTU RGB+D dataset. Then the three last layers in the network have been fine-tuned using a cyclical learning rate with the same schedule that for the NTU dataset. Besides, since the images provided in the NTU RGB+D dataset are masked, there has been necessary remove the background for the images belonging to these two datasets.

For the evaluation or the proposal, there have been used the evaluation protocols proposed for the authors of each dataset. Furthermore, the obtained results are compared to other works evaluated in these datasets. It is worth highlighting that most of the works that are based on DNNs use the large scale NTU RGB+D dataset because of the number of available samples, whereas the works based on classical methods are usually evaluated in NWUCLA or UWA3DII.


\subsection{Model performance evaluation }
\subsection{Experiments on NTU RGB+D dataset}
The proposed evaluations in~\cite{Shahroudy_2016_CVPR} for separating data between training and test for the NTU RBG+D dataset, that have been explained in the previous section, have been employed. 

The  method proposed in this paper has achieved an accuracy on NTU RGB+D dataset of 78.13\% (CS) and 80.31\% (CV) maintaining a very low computational cost, as detailed below, where an analysis of the different per-classes recognition accuracies, as well as a comparison with previous methods in terms of accuracy and computational cost are presented.

\begin{figure}[!htbp]
	\centering
	\includegraphics[width=0.9\linewidth]{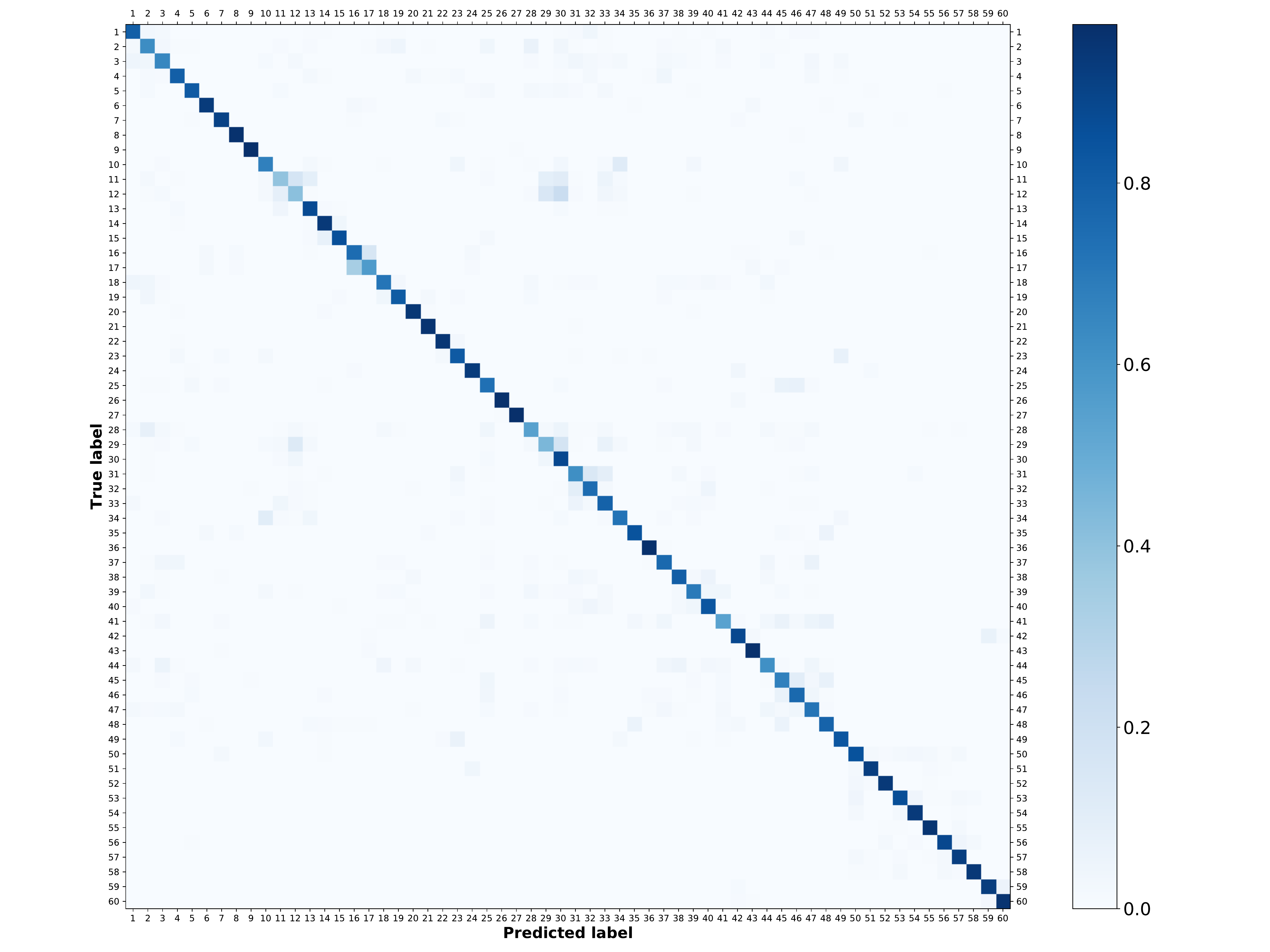}
	\caption{Confusion matrix of test results for the 3DFCNN on NTU RGB+D with 60 human actions (CV evaluation). Action indexes in figure correspond to the indexes used in author's webpage~\cite{ntuweb}.}
	\label{fig_confusion-matrix}
\end{figure}

The confusion matrix obtained for the NTU RGB+D dataset is shown in Fig.~\ref{fig_confusion-matrix}. It gives a general view of the model performance (for the CV evaluation), showing that it provides a good classification performance for the 60 different classes. 

In Table~\ref{Tabla:2} there are shown the 10 best recognized actions and the 10 most confused action pairs. Here, it can be seen that the proposed model can not totally recognize very similar actions like \emph{reading} (mostly confused with \emph{writing}), \emph{writing} (with \emph{type on a keyboard}) or \emph{play with phone/tablet} (with \emph{type on a keyboard}). It is noteworthy that, in general, the proposed model performs really well with actions where small objects are not involved. Nevertheless, where there are small objects which are discriminatory for the action recognition, e.g. phone/tablet, shoes, meals, toothbrush, etc., the model tends to missclassify towards a similar action. The reason for this may be the small input image size that is fed to the neural network ($64\times64$), along with the absence of color and texture of objects.

\begin{table}[!htbp]
	\caption{Top 10 accurate actions and confused pairs for the proposed model, including accuracy recognition per action.}
	\label{Tabla:2}
	\centering
	\small
	\begin{tabular}{| c |c|c|c |}
		
		\hline
		\multicolumn{2}{|c|}{\bfseries Top 10 recognized actions} & \multicolumn{2}{c|}{\bfseries Top 10 confused actions$^{*}$} \\
		\hline
		\hline
		1) Stand up & ($97.47\%$) & 1) Reading $\rightarrow$ Writing & ($39.56\%$)  \\
		2) Jump up & ($97.15\%$) & 2) Writing $\rightarrow$ Type on a keyboard & ($40.82\%$)  \\
		3) Hopping & ($96.84\%$) & 3) Play with phone/tablet $\rightarrow$ Type on a keyboard & ($44.62\%$)  \\
		4) Shake head & ($96.84\%$) & 4)Sneeze/cough $\rightarrow$ Nausea/vomiting & ($53.80\%$)  \\
		5) Falling down & ($96.84\%$) & 5) Phone call $\rightarrow$ Eat meal & ($54.43\%$)  \\
		6) Sit down & ($96.52\%$) & 6) Take off a shoe $\rightarrow$ Put on a shoe & ($56.96\%$)  \\
		7) Take off a hat/cap & ($95.87\%$) & 7) Headache $\rightarrow$ Brush teeth & ($61.39\%$) \\
		8) Walking apart & ($95.57\%$) & 8) Point to something $\rightarrow$ Taking a selfie & ($61.71\%$) \\
		9) Hugging & ($95.25\%$) & 9) Eat meal $\rightarrow$ Phone call & ($62.66\%$) \\
		10) Cheer up & ($94.94\%$) & 10) Brush teeth $\rightarrow$ Drink water & ($64.87\%$) \\
		\hline
		\multicolumn{2}{|c|}{\textbf{Average accuracy = $96.33\%$}} & \multicolumn{2}{c|}{\textbf{Average accuracy = $54.08\%$}} \\
		\hline
	\end{tabular}
	\begin{flushleft}\small{$^{*}$Numbers between parenthesis are the recognition accuracy of true action (before the arrow). }\end{flushleft}
\end{table}

\begin{figure}[!htbp]
	\centering
	\includegraphics[width=0.75\linewidth]{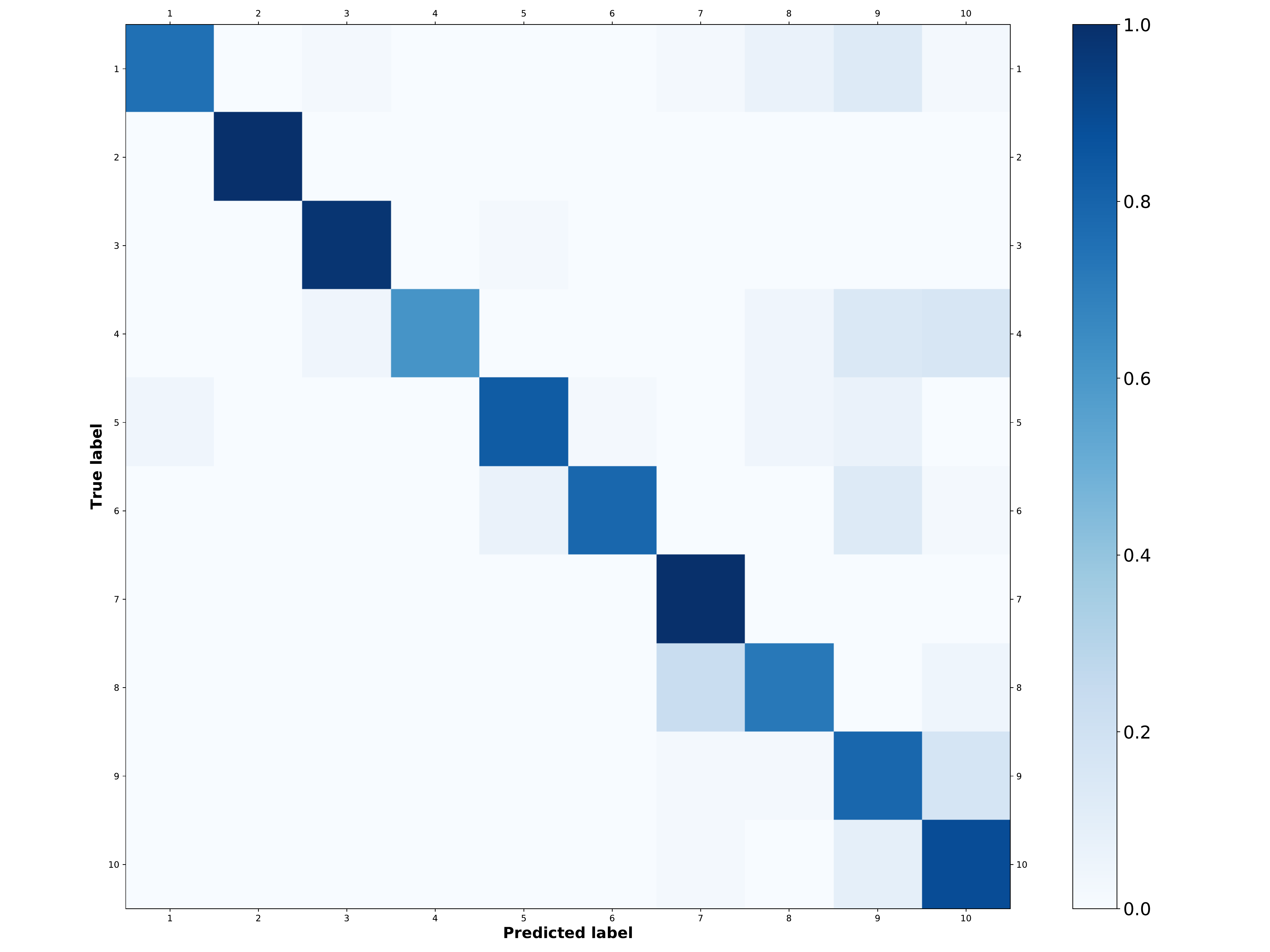}
	\caption{Confusion matrix of test results for the 3DFCNN on NWUCLA dataset}
	\label{fig:conf-nwucla}
\end{figure}

\begin{figure}[!htbp]
	\centering
	\includegraphics[width=0.75\linewidth]{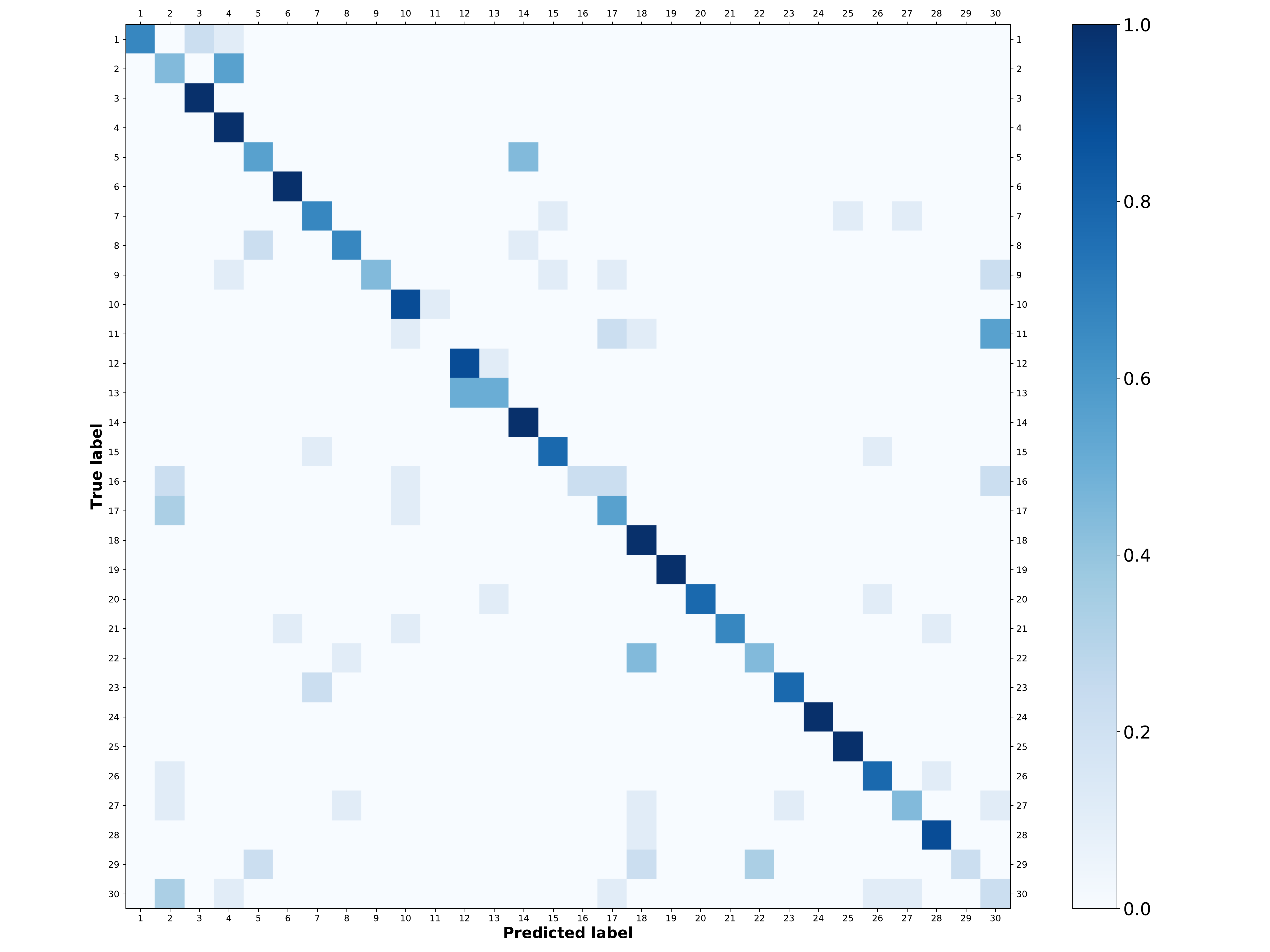}
	\caption{Confusion matrix of test results for the 3DFCNN on UWA3DII dataset}
	\label{fig:conf-uwa3dii}
\end{figure}

The confusion matrices have also been obtained for NWUCLA and UWA3DII datasets which, as it has been explained, include a lower number of actions. The results are shown in figure~\ref{fig:conf-nwucla} and~\ref{fig:conf-uwa3dii} respectively. As it can be seen in the figures, 3DFCNN is able to recognize action with a high accuracy for both datasets. 

\subsection{Comparison with state-of-the-art methods}

The three datasets described above have also been used to compare the method proposed in this work with other alternatives in the state of the art. Firstly, the 3DFCNN has been trained and evaluated with the large scale NTU RGB+D dataset, widely used by deep learning based methods due to the high number of samples available. In that database, the proposal presented has been compared 

To test the generalization capacity of the network, as well as its ability to adapt to other environments, it has been evaluated in two other image datasets that include depth information: NWUCLA and UWA3DII. These are two datasets with a much lower number of samples, which makes the training of DNNs difficult (so they are mainly used for the evaluation of proposals based on classical techniques). Furthermore, the used sensor is the Kinect V1, whose measurements contain more noise than the Kinect V2 camera, due to its different depth acquisition technology. 

Below there are first presented the results obtained with the NTU RGB+D dataset, which are compared to different with several state-of-the-art methods based on deep learning. Then, there are shown the results obtained with NWUCLA and UWA3DII. Since the number of samples is to small for most of the deep learning based approaches, the results in this section are compared to several state-of-the-art methods based on classical techniques.


\subsubsection{Comparison with NTU RGB+D dataset}


Table~\ref{tab:comparison-results} presents the results obtained in the NTU RGB+D dataset by different proposals in the literature. The results are divided in two different sections: on the top of the table, there are shown the results obtained for methods that use 3D skeletons, whereas on the bottom there are presented the approaches that only uses depth data (like the proposal in this paper). It is worth highlighting  that all the methods in table~\ref{tab:comparison-results} are based on DNNs.

The 3DFCNN proposed method in this paper has achieved an accuracy of 78.13\% (CS) and 80.37\% (CV) on NTU RGB+D dataset. As shown in Table~\ref{tab:comparison-results}, the 3DFCNN cannot achieve a recognition accuracy as high as other methods like, for instance, the one proposed in~\cite{xiao2019action}, but still it is a remarkable performance considering the challenging character of the NTU RGB+D dataset. 

\begin{table}[!htbp]
	\caption{Total average accuracy (\%) from different methods on NTU RGB+D dataset.}
	\label{tab:comparison-results}
	\centering
	\begin{tabular}{|c|c|c|}
		\hline
		\bfseries Method & \bfseries CS & \bfseries CV \\
		\hline\hline
		\bfseries Modality: 3D Skeleton & & \\
		ST-LSTM + Trust Gate (2016)~\cite{Shahroudy_2016_CVPR} & 69.2 & 77.7 \\
		Clips + CNN + MTLN (2017)~\cite{ke2017new} & 79.57 & 84.83 \\
		AGC-LSTM (2019)~\cite{si2019attention} & \textbf{89.2} & \textbf{95.0} \\
		\hline\hline
		\bfseries Modality: Depth & & \\
		DDI+DDNI+DDMNI (2018)~\cite{wang2018depth} & \textbf{87.08} & 84.22 \\
		HDDPDI (2019)~\cite{wu2019hierarchical} & 82.43 & \textbf{87.56} \\
		Multi-view dynamic images+CNN (2019)~\cite{xiao2019action} & 84.6 & 87.3 \\
		\hline
		Proposed method (3DFCNN) & 78.13 & 80.37 \\
		\hline
	\end{tabular}
\end{table}

Moreover, the performance of the proposed method gets more valuable when it is taken into account its simplicity and the absence of a costly pre-processing. This last characteristic allows the method to be quite fast when it is compared with more complex methods of the state-of-the-art. Table~\ref{tab:comparison-times} shows the computational cost in terms of average time consumption per video in test phase of some previous methods. The first four computational cost results shown were published in~\cite{wang2018depth}, computed with a different hardware than in the present work. Even so, it can give an idea of the reduced computational cost of our method compared, for instance, with the method which uses three sets of dynamic images (DDI+DDNI+DDMNI~\cite{wang2018depth}) which achieves a better recognition rate on NTU RGB+D dataset. The 3DFCNN model presented in this paper achieves the lowest time consumption, with an average time per 30-frames video sequence of $0.09$ s. This value has been computed using a NVIDIA GeForce GTX 1080 with 8 GB, through a set of 10\,000 randomly-chosen video samples from the NTU RGB+D dataset. The same GPU is used in~\cite{xiao2019action}, where multi-view dynamic images were used. In that paper, they also used a Intel(R) Xeon(R) E5-2630 V3 CPU running at 2.4 GHz to generate the multi-view dynamic images, which was the main source of computational cost, yielding a total average time consumption of 51.02 s per video. Therefore, this high computational cost prevent this method to be used in real time applications like, e.g., video surveillance and health care. However, the proposed method has both a sufficient high action recognition accuracy and yet a very low computational cost.

\begin{table}[!t]
	\renewcommand{\arraystretch}{1.3}
	\caption{Time consumption comparison of some action recognition methods with available data. The time values are the average of several time consumption values from different video samples of the dataset. See the text for more details. }
	\label{tab:comparison-times}
	\centering
	\begin{tabular}{|l|c|}
		\hline
		\bfseries Method & \bfseries Time (s) \\ 
		\hline \hline
		MSFK+DeepID~\cite{wan2015explore}					 & 41.00 \\
		SFAM~\cite{wang2017scene}							 &  6.33 \\
		WHDMM~\cite{wang2015action} 						 &  0.62 \\
		DDI+DDNI+DDMNI~\cite{wang2018depth} 				 & 62.03 \\
		Multi-view dynamic images+CNN~\cite{xiao2019action} & 51.02 \\
		\hline
		Proposed method (3DFCNN) 					& \textbf{ 0.09} \\
		\hline
	\end{tabular}
\end{table}

\subsubsection{Comparison with UWA3DII and NWUCLA datasets}

Below, there are shown the results obtained with UWA3DII (\ref{tab:exp-UWA3DII}) and NWUCLA (\ref{tab:exp-Northwestern-UCLA}) datasets. The results with both datasets are shown in the same section because of the similarities between their characteristics (type of sensor, number of people, number of samples, etc.). There is also presented the comparison to other three different proposals. It should be noted that most of the previous works that use these datasets are based in classical methods, since the reduced number of samples make it difficult training a DNN. On one hand, the first three works propose different new descriptors for action recognition: the Comparative Coding Descriptor (CCD)~\cite{cheng2012human},  the histogram of the surface normal orientation in the 4D space (HON4D)~\cite{oreifej2013hon4d}, and the Histogram of Oriented Principal Components (HOPC)~\cite{rahmani2014hopc}. On the other hand, the proposal in~\cite{xiao2019action} is based on a CNN combined to dynamic images, but it requires a costly pre-processing to obtain the dynamic images.

\begin{table}[htbp]
	\caption{Recognition accuracy (\%) for different depth maps-based methods 
		(CCD~\cite{cheng2012human}, HON4D~\cite{oreifej2013hon4d}, HOPC~\cite{uwa3dii} and 
		MVDI~\cite{xiao2019action}) on the UWA3DII dataset. 
		The evaluation criteria for this dataset consists of using two camera views for training and 
		the other two for testing.}
	\label{tab:exp-UWA3DII}
	\centering
	\resizebox{\textwidth}{!}{%
		\begin{tabular}{|l|cc|cc|cc|cc|cc|cc|c|}
			\hline
			Train & \multicolumn{2}{c|}{V$_{1}$+V$_{2}$} & 
			\multicolumn{2}{c|}{V$_{1}$+V$_{3}$} & 
			\multicolumn{2}{c|}{V$_{1}$+V$_{4}$} & 
			\multicolumn{2}{c|}{V$_{2}$+V$_{3}$} & 
			\multicolumn{2}{c|}{V$_{2}$+V$_{4}$} & 
			\multicolumn{2}{c|}{V$_{3}$+V$_{4}$} & 
			\multirow{2}{*}{Mean} \\
			Test & V$_{3}$ & V$_{4}$ & V$_{2}$ & V$_{4}$ & V$_{2}$ & V$_{3}$ & 
			V$_{1}$ & V$_{4}$ & V$_{1}$ & V$_{3}$ & V$_{1}$ & V$_{2}$ & \\
			\hline \hline
			CCD 	& 10.5 & 13.6 & 10.3 & 12.8 & 11.1 &  8.3 & 
			10.0 & 7.7  & 13.1 & 13.0 & 12.9 & 10.8 & 11.2 \\
			HON4D 	& 31.1 & 23.0 & 21.9 & 10.0 & 36.6 & 32.6 & 
			47.0 & 22.7 & 36.6 & 16.5 & 41.4 & 26.8 & 28.9 \\
			HOPC 	& 52.7 & 51.8 & 59.0 & \textbf{57.5} & 42.8 & 44.2 & 58.1 & 
			38.4 & 63.2 & 43.8 & 66.3 & 48.0 & 52.2 \\
			MVDI 	& \textbf{77.0} & \textbf{59.5} & \textbf{68.3} & 57.2 & 57.8 & 
			\textbf{72.9} & \textbf{80.3} & \textbf{51.3} & \textbf{76.6} &  
			\textbf{69.5} & \textbf{78.8} & 67.9 & 68.1 \\
			\hline
			3DFCNN 	& 68.3 & 54.6 & 66.7 & 51.5 & \textbf{68.2} & 67.2 & 74.3 & 49.7 & 75.1 & 61.9 & 73.5 & \textbf{88.3} & 66.6 \\
			\hline
	\end{tabular}}
	
\end{table}


\begin{table}[htbp]
	\caption{Recognition accuracy (\%) for different depth maps-based methods on the Northwestern-UCLA dataset. The evaluation criteria for this dataset consists of using the first two camera views for training and the remaining view for testing.}
	\label{tab:exp-Northwestern-UCLA}
	\centering
	\begin{tabular}{|l|c|}
		\hline
		\bfseries Method & Acc.\\
		\hline
		\hline
		CCD~\cite{cheng2012human} & 34.4 \\
		HON4D~\cite{oreifej2013hon4d} & 39.9 \\
		HOPC~\cite{uwa3dii} & 80.0 \\
		MVDI~\cite{xiao2019action} & \bfseries 84.2 \\
		\hline
		Proposed method (3DFCNN) & 83.6 \\
		\hline
	\end{tabular}
	
\end{table}

As it can be seen in tables~\ref{tab:exp-UWA3DII} and~\ref{tab:exp-Northwestern-UCLA}, despite the small number of samples for fine-tuning in these datasets, the results of the 3DFCNN proposed in this paper outperforms the approaches based on classical methods. Besides, the results are close to those obtained using the proposal in~\cite{xiao2019action}, surpassing them in some cases. These results are even more significant considering the reduced computational cost of the proposal compared to~\cite{xiao2019action}.



\section{Conclusions}
\label{sec:conclusions}

An end-to-end trainable deep learning approach for action recognition from depth videos is proposed. The model is a 3D fully convolutional neural network, named 3DFCNN, which automatically encodes spatial and temporal patterns of depth sequences without any costly pre-processing. Furthermore, an efficient data generation system and a particular training strategy were proposed. Newly appeared deep learning techniques like learning rate range test, cyclical learning rate schedule and fully convolutional architecture were used in order to improve the model performance. An exhaustive experimental evaluation of the proposal has been carried out, using three different publicly available datasets. 
Experimental results on the large-scale NTU RGB+D dataset show the proposed method achieves action recognition accuracy close to state-of-the-art deep learning based methods, while drastically reducing the computational cost because of its relatively simple structure. This property would allow the proposed 3DFCNN model to run on real time applications, like video surveillance, health care services, video analysis and human-computer interaction. Besides, results within smaller NWUCLA and UWA3DII datasets show that the proposal reliability overtakes that of different methods based on classical computer vision techniques, and obtains results comparable to those from other state-of-the-art methods based on deep learning.

As most of the action recognition methods, the 3DFCNN model tends to confuse similar actions in which there are just small discriminatory objects or short motions, like within actions \emph{writing} and \emph{reading}. Improving recognition accuracy of such actions is still an open problem and constitutes a line of future work for the proposal here presented.

\section*{Acknowledgment}

This work has been supported by the Spanish Ministry of Economy and Competitiveness under projects HEIMDAL-UAH (TIN2016-75982-C2-1-R) and ARTEMISA (TIN2016-80939-R), and by the University of Alcalá under projects ACERCA (CCG2018/EXP-029) and ACUFANO (CCG19/IA-024).

Portions of the research in this paper used the ``NTU RGB+D (or NTU RGB+D 120) Action Recognition Dataset" made available by the ROSE Lab at the Nanyang Technological University, Singapore.

\bibliographystyle{ieee}
\bibliography{3DFCNN}

\begin{thebibliography}{10}\itemsep=-1pt

\bibitem{ntuweb}
{NTU RGB+D Action Recognition dataset}, 2016.
\newblock Available online:
  \url{http://rose1.ntu.edu.sg/datasets/actionrecognition.asp} (Last access
  12/11/2019).

\bibitem{al2018human}
R.~Al-Akam, D.~Paulus, and D.~Gharabaghi.
\newblock Human action recognition based on 3d convolution neural networks from
  rgbd videos.
\newblock 2018.

\bibitem{ashraf2014}
N.~Ashraf, C.~Sun, and H.~Foroosh.
\newblock View invariant action recognition using projective depth.
\newblock {\em Computer Vision and Image Understanding}, 123:41--52, 2014.

\bibitem{Baptista2016}
M.~{Baptista-Ríos}, C.~{Martínez-García}, C.~{Losada-Gutiérrez}, and
  M.~{Marrón-Romera}.
\newblock Human activity monitoring for falling detection. a realistic
  framework.
\newblock In {\em 2016 International Conference on Indoor Positioning and
  Indoor Navigation (IPIN)}, pages 1--7, Oct 2016.

\bibitem{chaquet2013}
J.~M. Chaquet, E.~J. Carmona, and A.~Fernández-Caballero.
\newblock A survey of video datasets for human action and activity recognition.
\newblock {\em Computer Vision and Image Understanding}, 117(6):633 -- 659,
  2013.

\bibitem{chen2017}
C.~Chen, R.~Jafari, and N.~Kehtarnavaz.
\newblock A survey of depth and inertial sensor fusion for human action
  recognition.
\newblock {\em Multimedia Tools and Applications}, 76(3):4405--4425, 2017.

\bibitem{chen2016real}
C.~Chen, K.~Liu, and N.~Kehtarnavaz.
\newblock Real-time human action recognition based on depth motion maps.
\newblock {\em Journal of real-time image processing}, 12(1):155--163, 2016.

\bibitem{cheng2012human}
Z.~Cheng, L.~Qin, Y.~Ye, Q.~Huang, and Q.~Tian.
\newblock Human daily action analysis with multi-view and color-depth data.
\newblock In {\em European Conference on Computer Vision}, pages 52--61.
  Springer, 2012.

\bibitem{chou2018robust}
K.-P. Chou, M.~Prasad, D.~Wu, N.~Sharma, D.-L. Li, Y.-F. Lin, M.~Blumenstein,
  W.-C. Lin, and C.-T. Lin.
\newblock Robust feature-based automated multi-view human action recognition
  system.
\newblock {\em IEEE Access}, 6:15283--15296, 2018.

\bibitem{das2019}
S.~Das, M.~Thonnat, K.~Sakhalkar, M.~Koperski, F.~Bremond, and G.~Francesca.
\newblock A new hybrid architecture for human activity recognition from rgb-d
  videos.
\newblock In I.~Kompatsiaris, B.~Huet, V.~Mezaris, C.~Gurrin, W.-H. Cheng, and
  S.~Vrochidis, editors, {\em MultiMedia Modeling}, pages 493--505, Cham, 2019.
  Springer International Publishing.

\bibitem{dawar2017real}
N.~Dawar, C.~Chen, R.~Jafari, and N.~Kehtarnavaz.
\newblock Real-time continuous action detection and recognition using depth
  images and inertial signals.
\newblock In {\em 2017 IEEE 26th International Symposium on Industrial
  Electronics (ISIE)}, pages 1342--1347. IEEE, 2017.

\bibitem{dipakkr}
Dipakkr.
\newblock 3d-cnn action recognition.
\newblock \url{https://github.com/dipakkr/3d-cnn-action-recognition}, 2018.

\bibitem{farooq2015survey}
A.~Farooq and C.~S. Won.
\newblock A survey of human action recognition approaches that use an rgb-d
  sensor.
\newblock {\em IEIE Transactions on Smart Processing \& Computing},
  4(4):281--290, 2015.

\bibitem{feichtenhofer2016}
C.~Feichtenhofer, A.~Pinz, and A.~Zisserman.
\newblock Convolutional two-stream network fusion for video action recognition.
\newblock In {\em Proceedings of the IEEE conference on computer vision and
  pattern recognition}, pages 1933--1941, 2016.

\bibitem{greff2016lstm}
K.~Greff, R.~K. Srivastava, J.~Koutn{\'\i}k, B.~R. Steunebrink, and
  J.~Schmidhuber.
\newblock Lstm: A search space odyssey.
\newblock {\em IEEE transactions on neural networks and learning systems},
  28(10):2222--2232, 2016.

\bibitem{han2013}
J.~Han, L.~Shao, D.~Xu, and J.~Shotton.
\newblock Enhanced computer vision with microsoft kinect sensor: A review.
\newblock {\em IEEE transactions on cybernetics}, 43(5):1318--1334, 2013.

\bibitem{hou2017spatially}
Y.~Hou, S.~Wang, P.~Wang, Z.~Gao, and W.~Li.
\newblock Spatially and temporally structured global to local aggregation of
  dynamic depth information for action recognition.
\newblock {\em IEEE Access}, 6:2206--2219, 2017.

\bibitem{hsu2016}
Y.-P. Hsu, C.~Liu, T.-Y. Chen, and L.-C. Fu.
\newblock Online view-invariant human action recognition using rgb-d
  spatio-temporal matrix.
\newblock {\em Pattern Recognition}, 60:215 -- 226, 2016.

\bibitem{Hu2015CVPR}
J.-F. Hu, W.-S. Zheng, J.~Lai, and J.~Zhang.
\newblock Jointly learning heterogeneous features for rgb-d activity
  recognition.
\newblock In {\em The IEEE Conference on Computer Vision and Pattern
  Recognition (CVPR)}, June 2015.

\bibitem{hu2018ECCV}
J.-F. Hu, W.-S. Zheng, J.~Pan, J.~Lai, and J.~Zhang.
\newblock Deep bilinear learning for rgb-d action recognition.
\newblock In {\em The European Conference on Computer Vision (ECCV)}, September
  2018.

\bibitem{ioffe2015batch}
S.~Ioffe and C.~Szegedy.
\newblock Batch normalization: Accelerating deep network training by reducing
  internal covariate shift.
\newblock {\em arXiv preprint arXiv:1502.03167}, 2015.

\bibitem{ji20123d}
S.~Ji, W.~Xu, M.~Yang, and K.~Yu.
\newblock 3d convolutional neural networks for human action recognition.
\newblock {\em IEEE transactions on pattern analysis and machine intelligence},
  35(1):221--231, 2012.

\bibitem{jozefowicz2015empirical}
R.~Jozefowicz, W.~Zaremba, and I.~Sutskever.
\newblock An empirical exploration of recurrent network architectures.
\newblock In {\em International Conference on Machine Learning}, pages
  2342--2350, 2015.

\bibitem{ke2017new}
Q.~Ke, M.~Bennamoun, S.~An, F.~Sohel, and F.~Boussaid.
\newblock A new representation of skeleton sequences for 3d action recognition.
\newblock In {\em Proceedings of the IEEE conference on computer vision and
  pattern recognition}, pages 3288--3297, 2017.

\bibitem{KHAIRE2018107}
P.~Khaire, P.~Kumar, and J.~Imran.
\newblock Combining cnn streams of rgb-d and skeletal data for human activity
  recognition.
\newblock {\em Pattern Recognition Letters}, 115:107 -- 116, 2018.
\newblock Multimodal Fusion for Pattern Recognition.

\bibitem{khurana2018deep}
R.~Khurana and A.~K.~S. Kushwaha.
\newblock Deep learning approaches for human activity recognition in video
  surveillance-a survey.
\newblock In {\em 2018 First International Conference on Secure Cyber Computing
  and Communication (ICSCCC)}, pages 542--544. IEEE, 2018.

\bibitem{kim2017interpretable}
T.~S. Kim and A.~Reiter.
\newblock Interpretable 3d human action analysis with temporal convolutional
  networks.
\newblock In {\em 2017 IEEE conference on computer vision and pattern
  recognition workshops (CVPRW)}, pages 1623--1631. IEEE, 2017.

\bibitem{Kingma2014}
D.~P. Kingma and J.~Ba.
\newblock Adam: {A} method for stochastic optimization.
\newblock {\em CoRR}, abs/1412.6980, 2014.

\bibitem{ko2018}
K.-E. Ko and K.-B. Sim.
\newblock Deep convolutional framework for abnormal behavior detection in a
  smart surveillance system.
\newblock {\em Engineering Applications of Artificial Intelligence}, 67:226 --
  234, 2018.

\bibitem{kong2019}
J.~Kong, T.~Liu, and M.~Jiang.
\newblock Collaborative multimodal feature learning for rgb-d action
  recognition.
\newblock {\em Journal of Visual Communication and Image Representation},
  59:537 -- 549, 2019.

\bibitem{imagenet2012}
A.~Krizhevsky, I.~Sutskever, and G.~E. Hinton.
\newblock Imagenet classification with deep convolutional neural networks.
\newblock In F.~Pereira, C.~J.~C. Burges, L.~Bottou, and K.~Q. Weinberger,
  editors, {\em Advances in Neural Information Processing Systems 25}, pages
  1097--1105. Curran Associates, Inc., 2012.

\bibitem{Lange2001}
R.~Lange and P.~Seitz.
\newblock Solid-state time-of-flight range camera.
\newblock {\em Quantum Electronics, IEEE Journal of}, 37(3):390--397, Mar 2001.

\bibitem{laraba20173d}
S.~Laraba, M.~Brahimi, J.~Tilmanne, and T.~Dutoit.
\newblock 3d skeleton-based action recognition by representing motion capture
  sequences as 2d-rgb images.
\newblock {\em Computer Animation and Virtual Worlds}, 28(3-4):e1782, 2017.

\bibitem{li2017skeleton}
C.~Li, Q.~Zhong, D.~Xie, and S.~Pu.
\newblock Skeleton-based action recognition with convolutional neural networks.
\newblock In {\em 2017 IEEE International Conference on Multimedia \& Expo
  Workshops (ICMEW)}, pages 597--600. IEEE, 2017.

\bibitem{li2018independently}
S.~Li, W.~Li, C.~Cook, C.~Zhu, and Y.~Gao.
\newblock Independently recurrent neural network (indrnn): Building a longer
  and deeper rnn.
\newblock In {\em Proceedings of the IEEE Conference on Computer Vision and
  Pattern Recognition}, pages 5457--5466, 2018.

\bibitem{LIU201574}
A.-A. Liu, W.-Z. Nie, Y.-T. Su, L.~Ma, T.~Hao, and Z.-X. Yang.
\newblock Coupled hidden conditional random fields for rgb-d human action
  recognition.
\newblock {\em Signal Processing}, 112:74 -- 82, 2015.
\newblock Signal Processing and Learning Methods for 3D Semantic Analysis.

\bibitem{liu2019rgbd}
B.~Liu, H.~Cai, Z.~Ju, and H.~Liu.
\newblock Rgb-d sensing based human action and interaction analysis: A survey.
\newblock {\em Pattern Recognition}, 94:1--12, 2019.

\bibitem{liu2018viewpoint}
J.~Liu, N.~Akhtar, and M.~Ajmal.
\newblock Viewpoint invariant action recognition using rgb-d videos.
\newblock {\em IEEE Access}, 6:70061--70071, 2018.

\bibitem{Liu_2019_NTURGBD120}
J.~Liu, A.~Shahroudy, M.~Perez, G.~Wang, L.-Y. Duan, and A.~C. Kot.
\newblock Ntu rgb+d 120: A large-scale benchmark for 3d human activity
  understanding.
\newblock {\em IEEE Transactions on Pattern Analysis and Machine Intelligence},
  2019.

\bibitem{liu2016spatio}
J.~Liu, A.~Shahroudy, D.~Xu, and G.~Wang.
\newblock Spatio-temporal lstm with trust gates for 3d human action
  recognition.
\newblock In {\em European Conference on Computer Vision}, pages 816--833.
  Springer, 2016.

\bibitem{liu2018t}
K.~Liu, W.~Liu, C.~Gan, M.~Tan, and H.~Ma.
\newblock T-c3d: Temporal convolutional 3d network for real-time action
  recognition.
\newblock In {\em Thirty-second AAAI conference on artificial intelligence},
  2018.

\bibitem{luo2017unsupervised}
Z.~Luo, B.~Peng, D.-A. Huang, A.~Alahi, and L.~Fei-Fei.
\newblock Unsupervised learning of long-term motion dynamics for videos.
\newblock In {\em Proceedings of the IEEE Conference on Computer Vision and
  Pattern Recognition}, pages 2203--2212, 2017.

\bibitem{maas2013rectifier}
A.~L. Maas, A.~Y. Hannun, and A.~Y. Ng.
\newblock Rectifier nonlinearities improve neural network acoustic models.
\newblock In {\em Proc. icml}, volume~30, page~3, 2013.

\bibitem{oreifej2013hon4d}
O.~Oreifej and Z.~Liu.
\newblock Hon4d: Histogram of oriented 4d normals for activity recognition from
  depth sequences.
\newblock In {\em Proceedings of the IEEE conference on computer vision and
  pattern recognition}, pages 716--723, 2013.

\bibitem{poppe2010}
R.~Poppe.
\newblock A survey on vision-based human action recognition.
\newblock {\em Image and vision computing}, 28(6):976--990, 2010.

\bibitem{uwa3dii}
H.~{Rahmani}, A.~{Mahmood}, D.~{Huynh}, and A.~{Mian}.
\newblock Histogram of oriented principal components for cross-view action
  recognition.
\newblock {\em IEEE Transactions on Pattern Analysis and Machine Intelligence},
  38(12):2430--2443, Dec 2016.

\bibitem{rahmani2014hopc}
H.~Rahmani, A.~Mahmood, D.~Q. Huynh, and A.~Mian.
\newblock Hopc: Histogram of oriented principal components of 3d pointclouds
  for action recognition.
\newblock In {\em European conference on computer vision}, pages 742--757.
  Springer, 2014.

\bibitem{sadanand2012}
S.~Sadanand and J.~J. Corso.
\newblock Action bank: A high-level representation of activity in video.
\newblock In {\em Computer Vision and Pattern Recognition (CVPR), 2012 IEEE
  Conference on}, pages 1234--1241. IEEE, 2012.

\bibitem{schindler2008action}
K.~Schindler and L.~Van~Gool.
\newblock Action snippets: How many frames does human action recognition
  require?
\newblock In {\em 2008 IEEE Conference on Computer Vision and Pattern
  Recognition}, pages 1--8. IEEE, 2008.

\bibitem{Sell2014}
J.~Sell and P.~O'Connor.
\newblock The {X}box one system on a chip and {K}inect sensor.
\newblock {\em Micro, IEEE}, 34(2):44--53, Mar 2014.

\bibitem{Shahroudy_2016_CVPR}
A.~Shahroudy, J.~Liu, T.-T. Ng, and G.~Wang.
\newblock Ntu rgb+d: A large scale dataset for 3d human activity analysis.
\newblock In {\em The IEEE Conference on Computer Vision and Pattern
  Recognition (CVPR)}, June 2016.

\bibitem{si2019attention}
C.~Si, W.~Chen, W.~Wang, L.~Wang, and T.~Tan.
\newblock An attention enhanced graph convolutional lstm network for
  skeleton-based action recognition.
\newblock In {\em Proceedings of the IEEE Conference on Computer Vision and
  Pattern Recognition}, pages 1227--1236, 2019.

\bibitem{chahramani2014}
K.~Simonyan and A.~Zisserman.
\newblock Two-stream convolutional networks for action recognition in videos.
\newblock In Z.~Ghahramani, M.~Welling, C.~Cortes, N.~D. Lawrence, and K.~Q.
  Weinberger, editors, {\em Advances in Neural Information Processing Systems
  27}, pages 568--576. Curran Associates, Inc., 2014.

\bibitem{simonyan2014deep}
K.~Simonyan and A.~Zisserman.
\newblock Very deep convolutional networks for large-scale image recognition,
  2014.

\bibitem{Singh2019}
T.~Singh and D.~K. Vishwakarma.
\newblock Human activity recognition in video benchmarks: A survey.
\newblock In B.~S. Rawat, A.~Trivedi, S.~Manhas, and V.~Karwal, editors, {\em
  Advances in Signal Processing and Communication}, pages 247--259, Singapore,
  2019. Springer Singapore.

\bibitem{smith2017cyclical}
L.~N. Smith.
\newblock Cyclical learning rates for training neural networks.
\newblock In {\em 2017 IEEE Winter Conference on Applications of Computer
  Vision (WACV)}, pages 464--472. IEEE, 2017.

\bibitem{song2017end}
S.~Song, C.~Lan, J.~Xing, W.~Zeng, and J.~Liu.
\newblock An end-to-end spatio-temporal attention model for human action
  recognition from skeleton data.
\newblock In {\em Thirty-first AAAI conference on artificial intelligence},
  2017.

\bibitem{Tafazzoli2010}
F.~Tafazzoli and R.~Safabakhsh.
\newblock Model-based human gait recognition using leg and arm movements.
\newblock {\em Engineering Applications of Artificial Intelligence}, 23(8):1237
  -- 1246, 2010.

\bibitem{wan2015explore}
J.~Wan, G.~Guo, and S.~Z. Li.
\newblock Explore efficient local features from rgb-d data for one-shot
  learning gesture recognition.
\newblock {\em IEEE transactions on pattern analysis and machine intelligence},
  38(8):1626--1639, 2015.

\bibitem{Wang2013ICCV}
H.~Wang and C.~Schmid.
\newblock Action recognition with improved trajectories.
\newblock In {\em The IEEE International Conference on Computer Vision (ICCV)},
  December 2013.

\bibitem{nwucla}
J.~{Wang}, X.~{Nie}, Y.~{Xia}, Y.~{Wu}, and S.~{Zhu}.
\newblock Cross-view action modeling, learning, and recognition.
\newblock In {\em 2014 IEEE Conference on Computer Vision and Pattern
  Recognition}, pages 2649--2656, June 2014.

\bibitem{wang2019generative}
L.~Wang, Z.~Ding, Z.~Tao, Y.~Liu, and Y.~Fu.
\newblock Generative multi-view human action recognition.
\newblock In {\em Proceedings of the IEEE International Conference on Computer
  Vision}, pages 6212--6221, 2019.

\bibitem{Wang2020}
L.~Wang, D.~Q. Huynh, and P.~Koniusz.
\newblock A comparative review of recent kinect-based action recognition
  algorithms.
\newblock {\em IEEE Transactions on Image Processing}, 29:15–28, 2020.

\bibitem{wang2016temporal}
L.~Wang, Y.~Xiong, Z.~Wang, Y.~Qiao, D.~Lin, X.~Tang, and L.~Van~Gool.
\newblock Temporal segment networks: Towards good practices for deep action
  recognition.
\newblock In {\em European conference on computer vision}, pages 20--36.
  Springer, 2016.

\bibitem{wang2018human}
L.~Wang, Y.~Xu, J.~Cheng, H.~Xia, J.~Yin, and J.~Wu.
\newblock Human action recognition by learning spatio-temporal features with
  deep neural networks.
\newblock {\em IEEE Access}, 6:17913--17922, 2018.

\bibitem{wang2018depth}
P.~Wang, W.~Li, Z.~Gao, C.~Tang, and P.~O. Ogunbona.
\newblock Depth pooling based large-scale 3-d action recognition with
  convolutional neural networks.
\newblock {\em IEEE Transactions on Multimedia}, 20(5):1051--1061, 2018.

\bibitem{wang2015convnets}
P.~Wang, W.~Li, Z.~Gao, C.~Tang, J.~Zhang, and P.~Ogunbona.
\newblock Convnets-based action recognition from depth maps through virtual
  cameras and pseudocoloring.
\newblock In {\em Proceedings of the 23rd ACM international conference on
  Multimedia}, pages 1119--1122. ACM, 2015.

\bibitem{wang2015deep}
P.~Wang, W.~Li, Z.~Gao, J.~Zhang, C.~Tang, and P.~Ogunbona.
\newblock Deep convolutional neural networks for action recognition using depth
  map sequences.
\newblock {\em arXiv preprint arXiv:1501.04686}, 2015.

\bibitem{wang2015action}
P.~Wang, W.~Li, Z.~Gao, J.~Zhang, C.~Tang, and P.~O. Ogunbona.
\newblock Action recognition from depth maps using deep convolutional neural
  networks.
\newblock {\em IEEE Transactions on Human-Machine Systems}, 46(4):498--509,
  2015.

\bibitem{wang2017scene}
P.~Wang, W.~Li, Z.~Gao, Y.~Zhang, C.~Tang, and P.~Ogunbona.
\newblock Scene flow to action map: A new representation for rgb-d based action
  recognition with convolutional neural networks.
\newblock In {\em Proceedings of the IEEE Conference on Computer Vision and
  Pattern Recognition}, pages 595--604, 2017.

\bibitem{wang2016large}
P.~Wang, W.~Li, S.~Liu, Z.~Gao, C.~Tang, and P.~Ogunbona.
\newblock Large-scale isolated gesture recognition using convolutional neural
  networks.
\newblock In {\em 2016 23rd International Conference on Pattern Recognition
  (ICPR)}, pages 7--12. IEEE, 2016.

\bibitem{wang2018survey}
P.~Wang, W.~Li, P.~Ogunbona, J.~Wan, and S.~Escalera.
\newblock Rgb-d-based human motion recognition with deep learning: A survey.
\newblock {\em Computer Vision and Image Understanding}, 171:118 -- 139, 2018.

\bibitem{wang2017structured}
P.~Wang, S.~Wang, Z.~Gao, Y.~Hou, and W.~Li.
\newblock Structured images for rgb-d action recognition.
\newblock In {\em Proceedings of the IEEE International Conference on Computer
  Vision}, pages 1005--1014, 2017.

\bibitem{weinland2011}
D.~Weinland, R.~Ronfard, and E.~Boyer.
\newblock A survey of vision-based methods for action representation,
  segmentation and recognition.
\newblock {\em Computer vision and image understanding}, 115(2):224--241, 2011.

\bibitem{weng2017spatio}
J.~Weng, C.~Weng, and J.~Yuan.
\newblock Spatio-temporal naive-bayes nearest-neighbor (st-nbnn) for
  skeleton-based action recognition.
\newblock In {\em Proceedings of the IEEE Conference on Computer Vision and
  Pattern Recognition}, pages 4171--4180, 2017.

\bibitem{wu2019hierarchical}
H.~Wu, X.~Ma, and Y.~Li.
\newblock Hierarchical dynamic depth projected difference images--based action
  recognition in videos with convolutional neural networks.
\newblock {\em International Journal of Advanced Robotic Systems},
  16(1):1729881418825093, 2019.

\bibitem{xiao2019action}
Y.~Xiao, J.~Chen, Y.~Wang, Z.~Cao, J.~T. Zhou, and X.~Bai.
\newblock Action recognition for depth video using multi-view dynamic images.
\newblock {\em Information Sciences}, 480:287--304, 2019.

\bibitem{xu2015empirical}
B.~Xu, N.~Wang, T.~Chen, and M.~Li.
\newblock Empirical evaluation of rectified activations in convolutional
  network.
\newblock {\em arXiv preprint arXiv:1505.00853}, 2015.

\bibitem{yan2018spatial}
S.~Yan, Y.~Xiong, and D.~Lin.
\newblock Spatial temporal graph convolutional networks for skeleton-based
  action recognition.
\newblock In {\em Thirty-Second AAAI Conference on Artificial Intelligence},
  2018.

\bibitem{zhang2016real}
B.~Zhang, L.~Wang, Z.~Wang, Y.~Qiao, and H.~Wang.
\newblock Real-time action recognition with enhanced motion vector cnns.
\newblock In {\em Proceedings of the IEEE conference on computer vision and
  pattern recognition}, pages 2718--2726, 2016.

\bibitem{zhang2019comprehensive}
H.-B. Zhang, Y.-X. Zhang, B.~Zhong, Q.~Lei, L.~Yang, J.-X. Du, and D.-S. Chen.
\newblock A comprehensive survey of vision-based human action recognition
  methods.
\newblock {\em Sensors}, 19(5):1005, 2019.

\bibitem{zhang2017}
J.~Zhang, Y.~Han, J.~Tang, Q.~Hu, and J.~Jiang.
\newblock Semi-supervised image-to-video adaptation for video action
  recognition.
\newblock {\em IEEE transactions on cybernetics}, 47(4):960--973, 2017.

\bibitem{zhang2016}
J.~Zhang, W.~Li, P.~O. Ogunbona, P.~Wang, and C.~Tang.
\newblock Rgb-d-based action recognition datasets: A survey.
\newblock {\em Pattern Recognition}, 60:86 -- 105, 2016.

\bibitem{zhang2017view}
P.~Zhang, C.~Lan, J.~Xing, W.~Zeng, J.~Xue, and N.~Zheng.
\newblock View adaptive recurrent neural networks for high performance human
  action recognition from skeleton data.
\newblock In {\em Proceedings of the IEEE International Conference on Computer
  Vision}, pages 2117--2126, 2017.

\bibitem{zhang2012microsoft}
Z.~Zhang.
\newblock Microsoft kinect sensor and its effect.
\newblock {\em IEEE multimedia}, 19(2):4--10, 2012.

\bibitem{zhao2012}
Y.~{Zhao}, Z.~{Liu}, L.~{Yang}, and H.~{Cheng}.
\newblock Combing rgb and depth map features for human activity recognition.
\newblock In {\em Proceedings of The 2012 Asia Pacific Signal and Information
  Processing Association Annual Summit and Conference}, pages 1--4, Dec 2012.

\end{thebibliography}

\end{document}